\documentclass[sigconf,nonacm]{acmart}
\AtBeginDocument{%
  \providecommand\BibTeX{{%
    \normalfont B\kern-0.5em{\scshape i\kern-0.25em b}\kern-0.8em\TeX}}}

\setcopyright{acmlicensed}
\copyrightyear{2018}
\acmYear{2018}
\acmDOI{XXXXXXX.XXXXXXX}

\acmConference[Conference acronym 'XX]{Make sure to enter the correct
  conference title from your rights confirmation emai}{June 03--05,
  2018}{Woodstock, NY}
\acmISBN{978-1-4503-XXXX-X/18/06}

\setcopyright{none}
\renewcommand\footnotetextcopyrightpermission[1]{} % removes footnote with conference information in first column
\usepackage{listings}
\usepackage{tcolorbox}
\usepackage{caption}
\usepackage{subcaption}
\usepackage{tabularx}
\usepackage{multirow}
\usepackage{bm}
\usepackage{framed}
\usepackage{hyperref}
\usepackage{xcolor}
\begin{document}

% \title{Bridging the Gap: Text Representation and Foundation Models for Mismatch Modality Generalizability}
\title{Towards Optimizing with Large Language Model}


\author{Pei-Fu Guo}
\authornote{Both authors contributed equally to this research.}
\affiliation{%
  \institution{National Taiwan University}
  % \streetaddress{1 Th{\o}rv{\"a}ld Circle}
  % \city{Hekla}
  \country{}
}
\email{r12922217@csie.ntu.edu.tw}

\author{Ying-Hsuan Chen}
\authornotemark[1]
\affiliation{%
  \institution{National Taiwan University}
  \country{}
}
\email{r12922044@csie.ntu.edu.tw}

\author{Yun-Da Tsai}
\affiliation{%
  \institution{National Taiwan University}
  \country{}
}
\email{f08946007@csie.ntu.edu.tw}

\author{Shou-De Lin}
\affiliation{%
 \institution{National Taiwan University}
 \country{}
}
\email{sdlin@csie.ntu.edu.tw}

% \affiliation{%

% \affiliation{%

% \affiliation{%

% \affiliation{%


\begin{abstract}
    In this study, we evaluate the optimization capabilities of Large Language Models (LLMs) across diverse mathematical and combinatorial optimization tasks, where each task is described in natural language. These tasks require LLM to iteratively generate and evaluate solutions through interactive prompting, where each optimization step involves generating new solutions based on past results and then pass to subsequent iterations. We demonstrate that LLMs can perform various optimization algorithms and act as effective black-box optimizers, capable of intelligently optimizing unknown functions. We also introduce three simple yet informative metrics to evaluate optimization performance, applicable across diverse tasks and less sensitive to test sample variations. Our findings reveal that LLMs excel at optimizing small-scale problems with limited data and their performance is significantly affected by the dimension of problem and values, highlighting the need for further research in LLM optimization.
\end{abstract}

\maketitle

\section{Introduction}
\label{sec:intro}

% intro
Large Language Models have demonstrated exceptional capabilities in reasoning across a variety of natural language-based tasks ~\cite{cot}. However, their potential extends beyond multiple-choice questions or single-question answering. This work explores LLMs’ effectiveness in optimization across diverse tasks and problem dimensions. Optimization involves iteratively generating and evaluating solutions to improve a given objective function. Our research assesses LLM performance in interactive optimization, where each step generates new solutions based on previous ones and their values.

We conduct our study with four different types of optimization algorithms: Gradient Descent, Hill Climbing, Grid Search, and Black Box Optimization. To provide a comprehensive evaluation of LLM performance, we introduce three distinct metrics. These metrics provide a multifaceted view of task performance and are applicable across a broad spectrum of optimization tasks, reducing sensitivity to sample variations.

Our findings suggest that LLMs show impressive optimization capabilities, especially in small-scale problems. However, their performance is notably affected by factors like sample size and value range. These observations underscore the need for further research within the domain of optimization tasks tailored for LLMs. It's important to note that our work does not aim to outperform state-of-the-art optimization algorithms for either mathematical optimization or combinatorial optimization problems. Instead, our goal is to showcase the potential of LLM in these optimization domains and find out limitations in these settings.

Our contributions are summarized as follows:
\begin{itemize}
    \item Exploring the potential of LLMs in mathematical and combinatorial optimization scenarios.
    \item Introduce three novel metrics for assessing LLM performance in optimization tasks.
    \item Delve into factors that influence LLM performance using our metrics, with a particular emphasis on the impact of problem dimension and task type.
\end{itemize}

The remainder of this paper is structured as follows. In Section~\ref{sec:related}, we present preliminary works on LLMs for addressing optimization challenges. In Section~\ref{sec:problem}, we defined 4 optimization algorithms in the case studies. In Section~\ref{sec:method}, we demonstrate that LLMs with iterative prompting strategy function as optimizers. In Section~\ref{sec:eval}, we present three metrics that we have designed to assess the overall performance of LLMs in undertaking optimization tasks. Section~\ref{sec:exp}, details our experimental results, showcasing the effectiveness of using LLMs as optimizers. In Section~\ref{sec:analysis}, we consolidated noteworthy observations and points of discussion from the experiments. Finally, Section~\ref{sec:conclude} summarizes and concludes the paper.

\section{Related Works}
\label{sec:related}
In various optimization scenarios, the utilization of Large Language Models (LLMs) has become indispensable for the development of optimization algorithms or agent systems capable of handling complex and informative text-based feedback. In this section, we summarize three significant related works that leverage LLMs to tackle optimization and reinforcement learning challenges. These works showcase the adaptability and effectiveness of LLMs in addressing optimization and learning challenges across various domains.

\textbf{Optimization by PROmpting (OPRO)}~\cite{llm_as_opt} OPRO harnesses LLMs as versatile optimizers by describing optimization tasks in natural language prompts. It iteratively generates and evaluates solutions from these prompts, demonstrating superior performance on tasks like linear regression and traveling salesman problems. OPRO outperforms human-designed prompts by up to 50\% on challenging tasks.

\textbf{Reflexion}~\cite{rl} Reflexion introduces a novel framework for training language agents that rely on linguistic feedback rather than traditional reinforcement learning. This framework delivers outstanding results, boasting a remarkable 91\% pass@1 accuracy on coding tasks—an exceptional 11\% improvement over previous state-of-the-art models. Reflexion's success underscores the potential of linguistic feedback as a powerful training mechanism.

\textbf{EvoPrompt}~\cite{pw} EvoPrompt automates prompt optimization by connecting LLMs with evolutionary algorithms. This automated process surpasses human-designed prompts by up to 25\% and outperforms existing automatic prompt generation methods by an impressive 14\%. EvoPrompt's success highlights the relationship between Large Language Models and traditional algorithms, showcasing the potential for enhanced problem-solving capabilities through this synergistic fusion.

\section{Problem setting}
\label{sec:problem}
We design four optimization tasks that require the model to algorithmically search for the optimal value of parameters. These tasks encompass Gradient-Descent, Hill-Climbing, Grid-Search, and Black-Box Optimization, each representing unique optimization domains: gradient-based, meta-heuristics, decision-theoretic, and Bayesian. In terms of parameter types, Grid-Search and Hill-Climbing involve discrete search spaces, while Gradient-Descent and Black-Box Optimization tackle continuous search spaces. Following is detailed information on each optimization task.

{\bfseries Gradient-Descent} assesses the model's proficiency in advanced calculations and its grasp of the principles of gradient descent. We instruct LLMs to undertake a conventional gradient descent optimization process based on the loss function they have defined. LLMs need to compute the gradient and update the parameters using the gradient information and the learning rate given.

{\bfseries Hill-Climbing} evaluate the LLM's capability to adhere to custom predefined rules they have not seen before. LLMs start with an initial solution and iteratively explore nearby solutions by making small incremental changes. In our task, neighboring solutions are generated by selecting a specific element within the solution and either increasing or decreasing it by one each time. Subsequently, the neighbor solution with the minimum loss is chosen as the new solution and passed to the next iteration.

{\bfseries Grid-Search} assesses the LLM's ability to conduct exhaustive searches and locate optimal solutions within a predefined search space. LLMs are tasked with generating all grid points and systematically searching for the point that results in the lowest loss according to the given loss function. 

{\bfseries Black-Box Optimization} evaluates the LLM's ability to make informed decisions and optimize in an abstract problem-solving context. We treat the LLMs as black boxes that try to fit an unknown loss function. We provide the LLM with a limited set of solutions, each paired with its respective true loss value. The LLM's objective is to discover new solutions that have lower losses than the existing solutions in each iteration by themselves.

\section{Methodologies}
\label{sec:method}
In this section, we show how LLMs, guided by iterative prompting, can effectively function as optimizers, akin to various optimization algorithms. To systematically navigate the search space, we introduce an iterative prompting framework that enables LLMs to incrementally achieve better solutions within the search space through iterative processes.

\begin{figure}[t]
    \centering
    \includegraphics[width=1.0\linewidth]{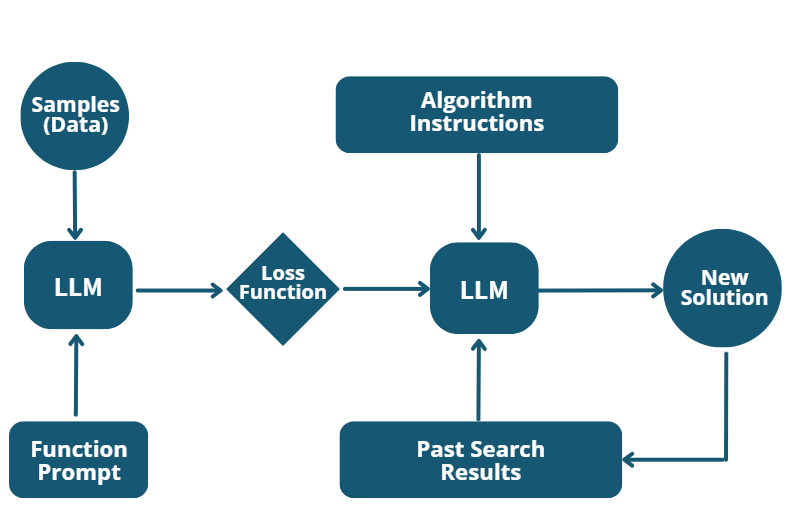}
    \caption{Overview of our prompting strategy. (1) LLMs formulate the loss function based on given samples. (2) Given algorithm instructions and past results, LLM generates a new solution.  (3) Calculate the loss of the new solution and add the solution-score pairs to the prompt of the next iteration. (4) Repeat the second and third steps until stop criteria are met.} 
    \label{FRAMEWORK}
\end{figure}

We applied Chain of Thoughts and iterative prompting as our prompting method. LLM will accomplish each step with reasoning thoughts as intermediate outputs. In each of these tasks (optimization algorithm), LLMs are initially required to formulate the loss function based on given samples. Then each optimization iteration is composed of two steps: (1) Generates new solution based on algorithm instructions and past search results (2) Calculate loss of new solution and add the results to the prompt of the next iteration. We keep repeating the two steps until the stop criteria are met. Figure~\ref{FRAMEWORK} shows an overview of how LLM performs optimization in interactive settings.

To create an interactive environment, we utilize the chat mode of GPTs, where the entire conversation history serves as the prompt. This allows LLMs to retain memory of past search results and reasoning paths. New instructions are appended to ongoing conversation records with each iteration. If the dialogue surpasses the token limit, earlier portions are removed. % Detailed meta prompt instructions and complete prompt examples for each optimization algorithm can be found in Appendix~\ref{app:prompt}.

\section{Evaluation}
\label{sec:eval}
We devised three novel metrics for the comprehensive evaluation of LLM capabilities. In this section, we will explain the design and objective of each metric. These metrics offer versatility in assessing LLM performance across diverse tasks, making concurrent evaluation easier. Their reliance on ratio measures, rather than differences, makes them less sensitive to sample variations.

\subsection{Goal Metric}
Goal metric evaluates how effectively LLMs perform optimization. It provides a quantitative measure of the degree to which the LLM contributes to minimizing the loss function values. In other words, ensuring that the ultimate solution loss is lower than the initial solution. We define the $goal$ $metric$ of a test sample $j$ as : 
\begin{equation}
        G_{j} = \frac{1}{N}\sum_{i = 1}^{N}\frac{loss_{LLM, init}-loss_{LLM, i}}{loss_{LLM, init}}
\end{equation}
where $loss_{LLM,init}$ is the initial solution loss of sample $j$, $loss_{LLM,i}$ is the LLM output loss of trial $i$, and $N$ is the number of trials per sample. The higher the metric value, the greater the progress in optimization. The goal metric plays a crucial role in our evaluation framework, particularly in scenarios where ground truth is absent, such as the Black-Box optimization scenarios. 

\subsection{Policy Metric}
Policy metric assesses the degree of alignment between the final model output and the ground truth. Beyond self-improvement, which is measured by $goal$ $metric$, it is also crucial to appraise the LLMs' capability to operate in a manner consistent with our truth model algorithm. This metric serves as an indicator of the LLM's adeptness in adhering to task-specific instructions. We define the $policy$ $metric$ of a test sample $j$ as : 
\begin{equation}
        P_{j} = \frac{1}{N}\sum_{i = 1}^{N}\frac{loss_{LLM,i}-loss_{truth}}{loss_{truth}}
\end{equation}
where $loss_{LLM,i}$ is the LLM output loss of trial $i$, $loss_{truth}$ is the ground truth of sample $j$ and $N$ is the number of trials. Since the policy metric measures the disparity between the ground truth and the LLM's output, a lower policy metric value indicates a more effective alignment of the LLM's actions with the prescribed guidelines. When the value is negative, it means that LLM's performance surpasses the ground truth.

\subsection{Uncertainty Metric}
Uncertainty metric quantifies the variability in the LLM's solutions under identical conditions. Stability is a crucial characteristic in optimization tasks. We hope that the LLMs produce identical results in every trial involving the same sample, even under conditions with temperatures greater than zero. We define the $uncertainty$ $metric$ of a test sample $j$ as : 
\begin{equation}
        U_{j} = \frac{1}{N}\sum_{i=1}^{N}(loss_{LLM, i}-\overline{loss_{LLM}})^{2}
\end{equation}
where $loss_{LLM, i}$ is the LLM output of the i-th trial, $\overline{loss_{LLM}}$ is the mean of the trial outputs and $N$ is the number of trials. A stable LLM can be more trusted for tasks that demand consistent and reproducible results. In our case, if the language model truly understands the context of problems, the final optimal output should be identical in every trial of the same sample.

\section{Experiments}
\label{sec:exp}
This section provides details of our experimental configurations and highlights the outcomes of experiments. Subsection ~\ref{sec:data} outlines the process of generating synthetic datasets for all optimization tasks, while subsection ~\ref{sec:detail} elucidates the detailed settings of our experiment. Lastly, subsection ~\ref{sec:result} offers a concise summary of the outcomes derived from our experiment.

\subsection{Dataset}
\label{sec:data}
In the experiment, we create five datasets with \(d\) values chosen from the set \(\{3, 6, 12, 24, 48\}\) and generate instances belonging to \([0, 10]^d\) in each dataset to examine sensitivity to the number of parameters, representing the dimension of the optimization problem. For instance, $d=3$ indicates that there are 3 variables in the loss function and the dimension of this optimization problem is 3. We then apply each instance to a loss function and find the true solution for each parameter search task. These authenticated solutions, coupled with their associated losses, not only serve as the ground truth for the tasks but also act as a pivotal benchmark against which the solutions derived by LLMs are systematically evaluated and compared in the ensuing analysis. %These true solutions and their losses serve as the ground truth for the tasks and are used as a benchmark to compare with the solutions found by Large Language Models afterward.

\subsection{Detailed Settings}
\label{sec:detail}
In our experiment, We set the LLM temperature to 0.8 and the reset as default. We performed 5 repetitions of the test for each instance in the dataset, with the LLM conducting 10 iterations of parameter search in each repetition. We excluded excessively biased results to prevent our metrics from being skewed by a minority of poorly performing test outcomes. All experiments employ the GPT-turbo-3.5 '0613' version as the Language Model.

\subsection{Main Results}
\label{sec:result}
We summarize the outcomes of our experiment and subsequently examine the common trends observed across all experiments. In every plot, the x-axis displays the dimension of the optimization problem. In the case of the goal metric and policy metric plots, the y-axis illustrates the average metric value for the respective tasks, while the shaded area in a lighter color delineates the confidence interval of the metric, denoted as $[value-std,  value+std]$. As for the uncertainty metric plot, the y-axis showcases the uncertainty metric value, which corresponds to the standard deviation of the LLM final solution loss. It is worth noting that the Goal Metric graph excludes the non-iterative Grid-Search task due to its non-iterative nature, while the Policy Metric graph omits the Black-Box task due to unattainable ground truth.

\begin{figure}[h]
      \begin{minipage}[h]{1.0\linewidth}
        \centering
        \includegraphics[width=1.0\textwidth]{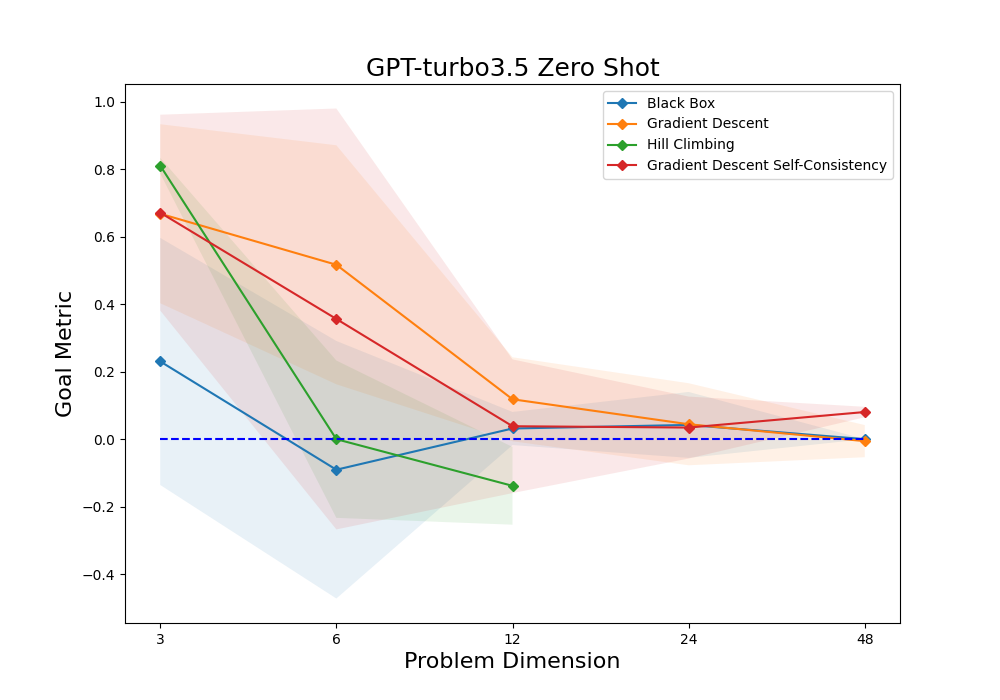}
        \label{fig:image1}
      \end{minipage}
      \hfill
      \begin{minipage}[h]{1.0\linewidth}
        \centering
        \includegraphics[width=1.0\textwidth]{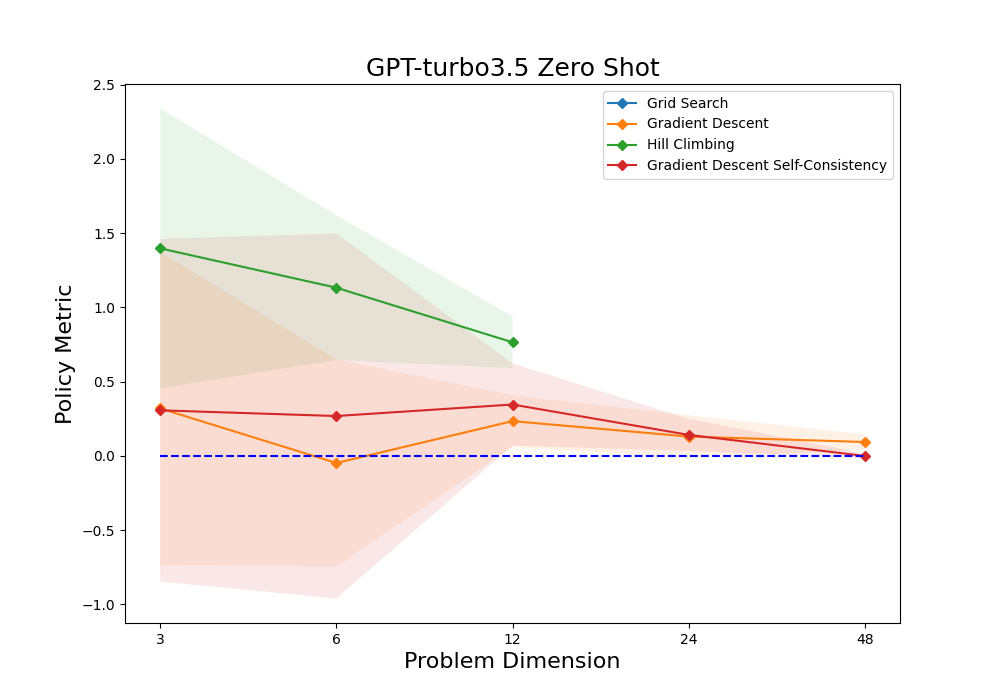}
        \label{fig:image2}
      \end{minipage}
      \caption{Goal Metric and Policy Metric hover from positive to near zero, signifying substantial optimization capability and alignment between LLM’s output and ground truth.}      
      \label{gpt-3.5-zeroshot}
\end{figure}

% A low Goal Metric and near-zero Policy Metric mean LLM performance matches the ground truth model. Conversely, high values for both metrics indicate optimization progress but poor alignment with instructions.
{\bfseries LLMs show strong optimization capabilities in small-scale problems.} Our experiments test the comprehensive optimization capabilities of LLMs. Observing figure~\ref{gpt-3.5-zeroshot}, GPT-turbo-3.5 showcases considerable optimization capabilities across various scenarios. Impressively, in the Gradient-Descent task, GPT-turbo-3.5 even surpasses the ground truth, particularly in the case of the sample dimension equal to six. It's also surprising that the model achieves respectable results in the Grid-Search task, considering it must compute a vast number of grid points, which increase exponentially as the dimension of the problem expands. The model faces challenges in the Hill-Climbing task, evident from a policy metric significantly exceeding zero. This suggests that meta-heuristics may pose greater difficulty for LLMs compared to other tasks.

\begin{figure}[h]
      \begin{minipage}[h]{1.0\linewidth}
        \centering
        \includegraphics[width=1.0\textwidth]{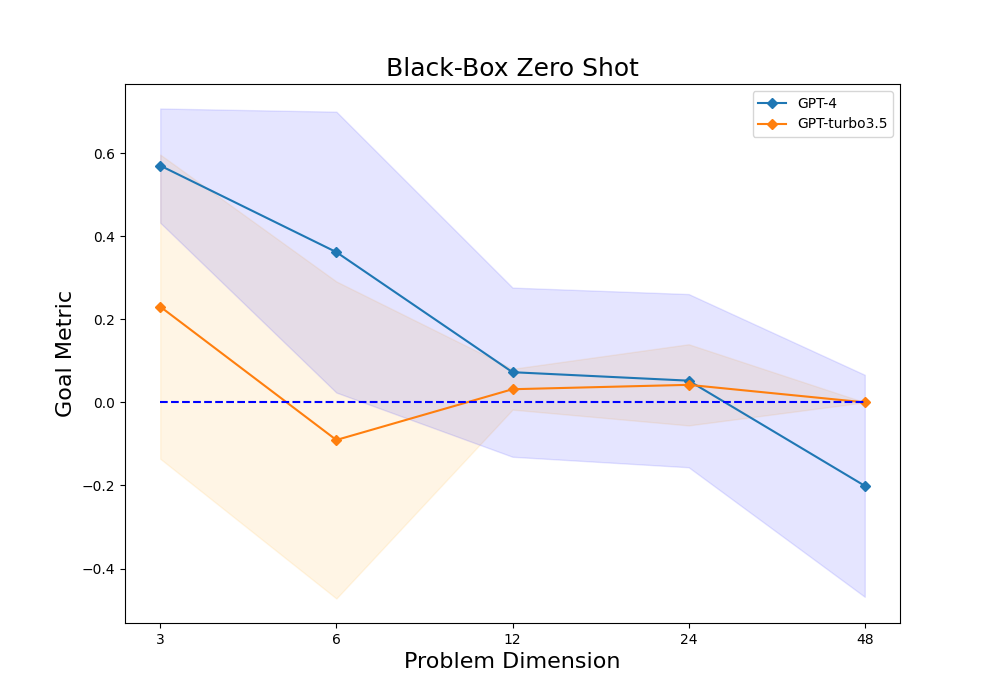}
        \label{fig:image3}
      \end{minipage}
      \hfill
      \begin{minipage}[h]{1.0\linewidth}
        \centering
        \includegraphics[width=1.0\textwidth]{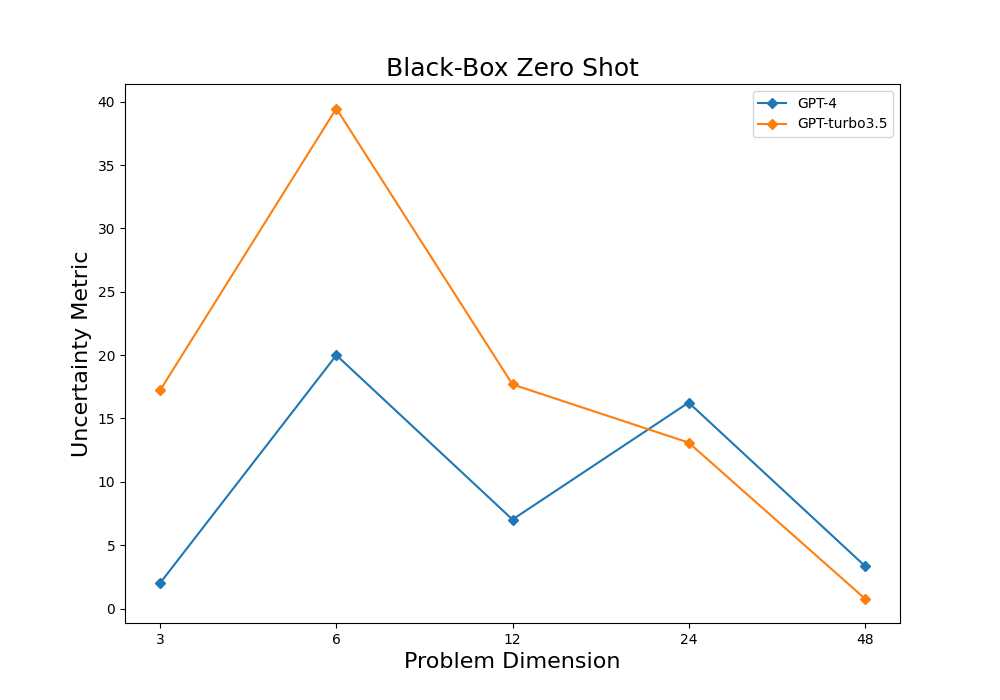}
        \label{fig:image4}
      \end{minipage}
      \caption{Goal Metric reflects the performance of LLMs as Black-Box optimizer, showing strong performance with instances of smaller dimensions.}
      \label{gpt-average-bb}
    \end{figure}

{\bfseries LLMs show potential as Black-Box Optimizers.} Favorable performance in Black-Box experiments suggests the use of LLM as an optimizer without giving any algorithm instructions. From figure~\ref{gpt-average-bb}, we can see that GPT-turbo-3.5 performs notably when the dimension of the problem is three, whereas GPT-4 excels when the dimensions are three and six. Interestingly, as the dimension increases, the performance of both models gradually diminishes. Eventually, GPT-4 edged out GPT-turbo-3.5 by a slight margin in optimization and stability.\\

\begin{figure}[h]
      \begin{minipage}[h]{1.0\linewidth}
        \centering
        \includegraphics[width=1.0\linewidth]{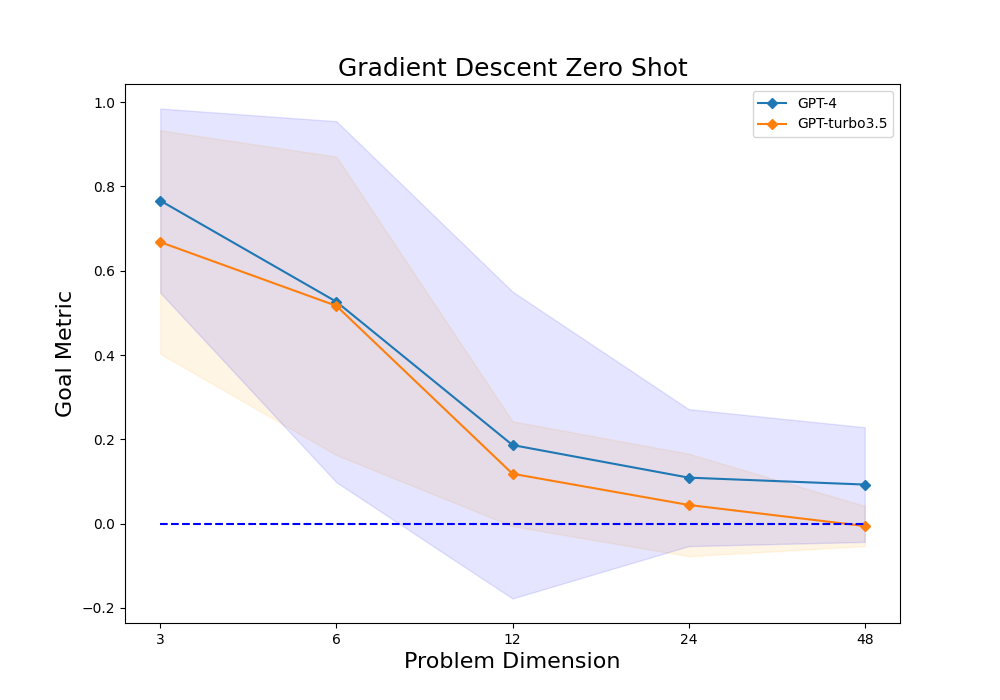}
        \label{fig:image5}
      \end{minipage}
      \hfill
      \begin{minipage}[h]{1.0\linewidth}
        \centering
        \includegraphics[width=1.0\linewidth]{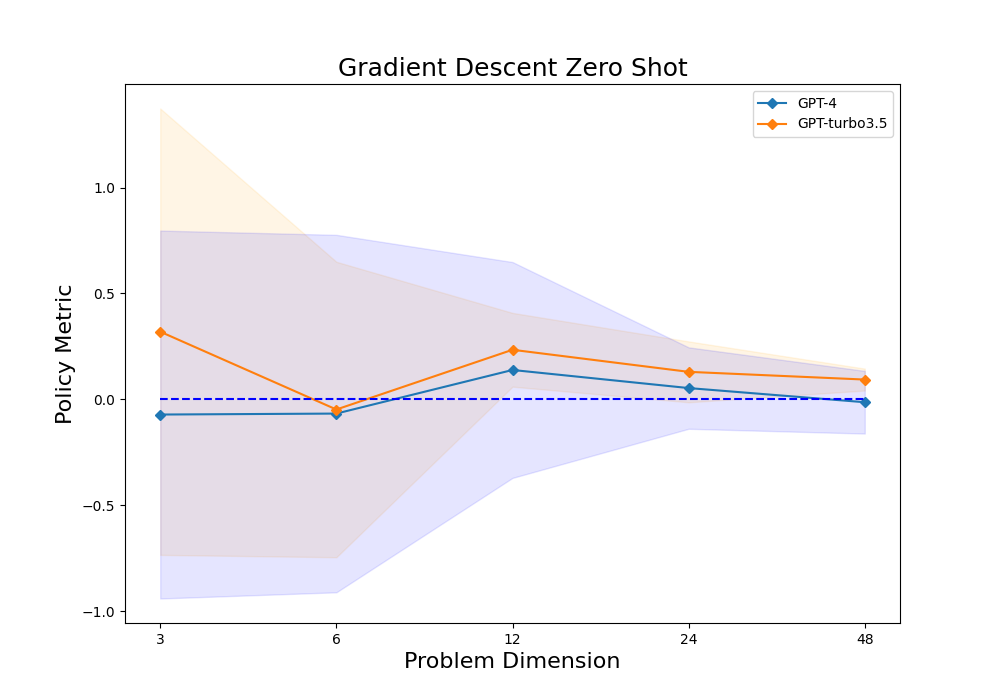}
        \label{fig:image6}
      \end{minipage}
      \caption{Low values in the Policy Metric and high positive values in the Goal Metric indicate the robust performance of the LLM in the gradient descent task.}
      \label{gpt-average-gd}
    \end{figure}

{\bfseries LLMs exhibit strong performance in Gradient-Descent.} Gradient-Descent experiment tests the model’s proficiency in advanced calculations and grasp of mathematics principles. Figure~\ref{gpt-average-gd} underscores this by revealing a policy metric that consistently hovers near zero, signifying a remarkable alignment between the LLM's output and the ground truth. Despite a decline in the goal metric as the sample size increases, the consistently low and stable value of the policy metric underscores the fact that GPT's performance in the gradient-descent task is nearly on par with the truth model.  

\section{Analysis and Discussion}
\label{sec:analysis}
In this section, we consolidate several crucial insights derived from our experimental results and subject them to analysis. 

% \subsection{Key Findings}
{\bfseries Pretrained Knowledge dominates the optimization capability of LLM.} Among all optimization tasks performed by LLMs, Gradient Descent emerges as the leading performer, while Hill-Climbing poses greater challenges. The main difference between the two tasks is that Hill-Climbing is a heuristic algorithm with more user-specific parameters, whereas gradient descent is an optimization algorithm that relies more on mathematical principles. This suggests that LLM optimization capabilities primarily stem from pretrained knowledge stored within the model parameters, rather than from context knowledge provided by users. Our findings align with previous research ~\cite{llm_knowledge, llm_summary, llm_aware} showing that language models often prioritize their prior knowledge over new context. Achieving balanced attention to both prior and context knowledge is essential for further research to improve the optimization capability of language models.
 
{\bfseries LLMs are potential hybrid optimizers.} The predominantly positive goal metric values across most tasks and datasets indicate LLMs' capability for optimization. This highlights their versatile capacity to optimize across different problem spaces, potentially allowing for the switching between optimization methods within a single task. Such switching can help LLMs better explore the solution space and escape local optima where they might get stuck. This is a significant advantage of LLMs in optimization, as they can easily change methods through a simple natural language prompt during iterations. Furthermore, LLMs can act as agents (world models) that use different algorithms as tools (actions), switching methods by evaluating the optimization path from past to present (state). This adaptability underscores the potential of LLMs to enhance optimization processes through dynamic method selection and strategic problem-solving.

\begin{figure}[h]
  \begin{minipage}[h]{1.0\linewidth}
    \centering
    \includegraphics[width=1.0\textwidth]{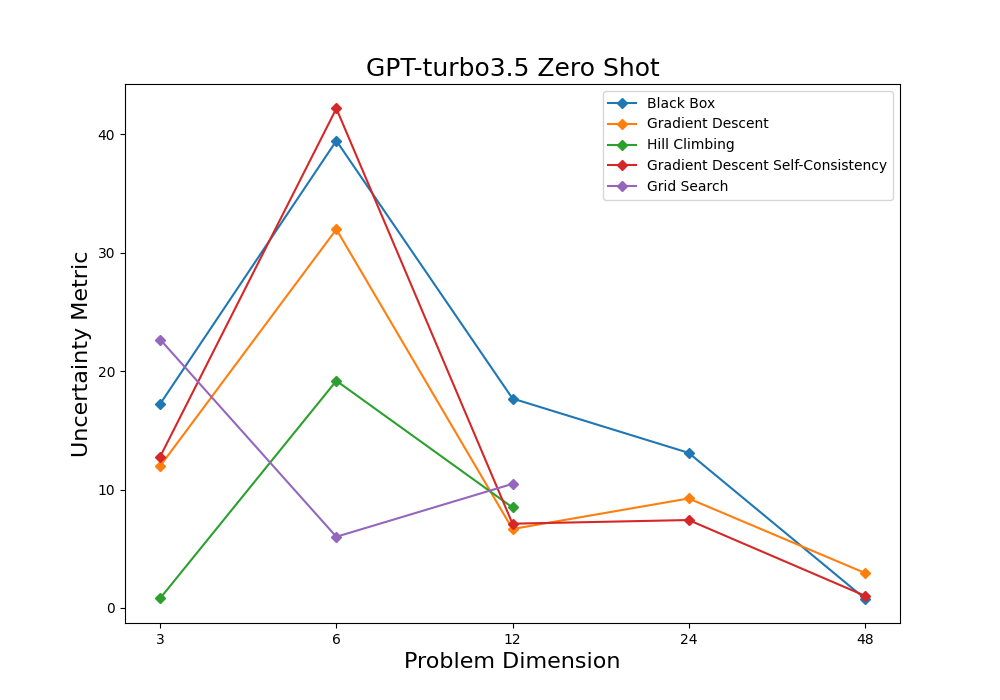}
    \label{fig:image7}
  \end{minipage}
  \hfill
  \begin{minipage}[h]{1.0\linewidth}
    \centering
    \includegraphics[width=\textwidth]{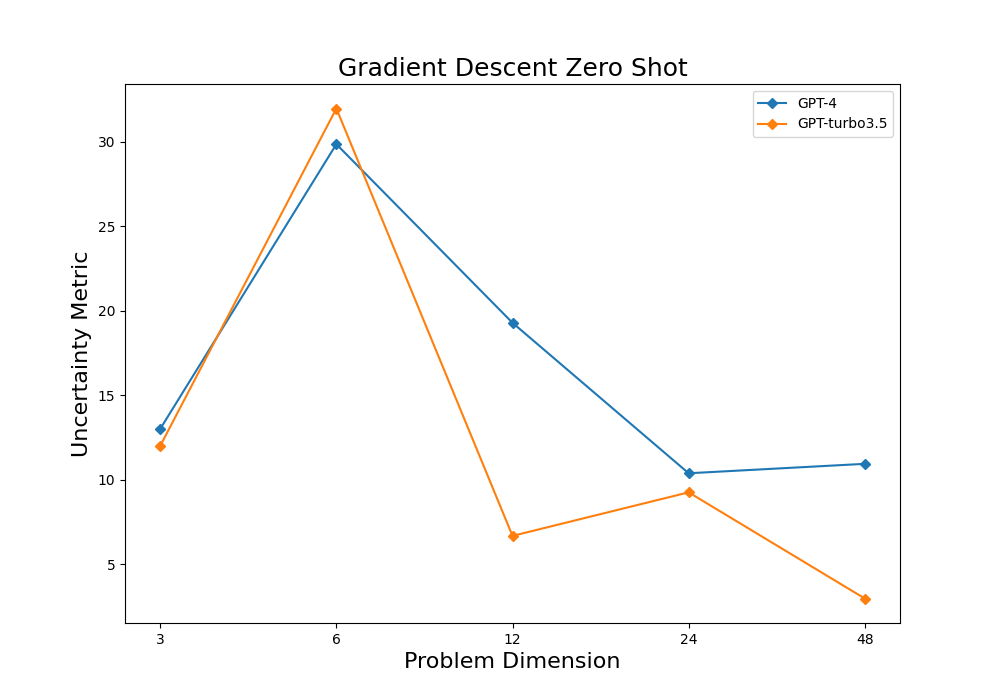}
    \label{fig:image8}
    \end{minipage}
\caption{An initial rise followed by a decline in the Uncertainty Metric with instance dimension growth suggests LLMs may have a richer sample space for small-scale problems, consistent across tasks and models.}

\label{uc}
\end{figure}

{\bfseries LLMs possess richer solution space in small-scale problems.} In our experiments, we observed high uncertainty metric values and significant variations in policy and goal metrics when samples had smaller dimensions. Interestingly, LLMs tend to perform more effectively with smaller dimension instances, suggesting a correlation between higher uncertainty and better performance. This consistent pattern across various tasks and models indicates that LLMs have a richer solution space when tackling small-scale problems. The expanded solution space leads to higher uncertainty, providing LLMs with a broader range of solutions to explore. This highlights the importance of dimension reduction in data preprocessing for effective optimization by LLMs. Figure~\ref{gpt-3.5-zeroshot} and~\ref{uc} both highlight the pattern of uncertainty, where the uncertainty initially rises and then gradually decreases. 

{\bfseries LLMs are sensitive to numerical values.} It's worth considering that the aforementioned results may be influenced by the inherent randomness in the generation of test samples. Previous research has indicated that LLMs may demonstrate preferences for particular numbers, words, and symbols~\cite{llm_rand}, which can introduce a level of bias in their responses. Given the high sensitivity of LLMs to the input prompt, the initial starting points and data provided can exert a significant influence on their outputs. In essence, the impact of instruction description and data initialization should be carefully considered when interpreting the results of LLM-based experiments to ensure a more accurate assessment of their performance.

\begin{figure}[h]
  \begin{minipage}[h]{1.0\linewidth}
    \centering
    \includegraphics[width=1.0\textwidth]{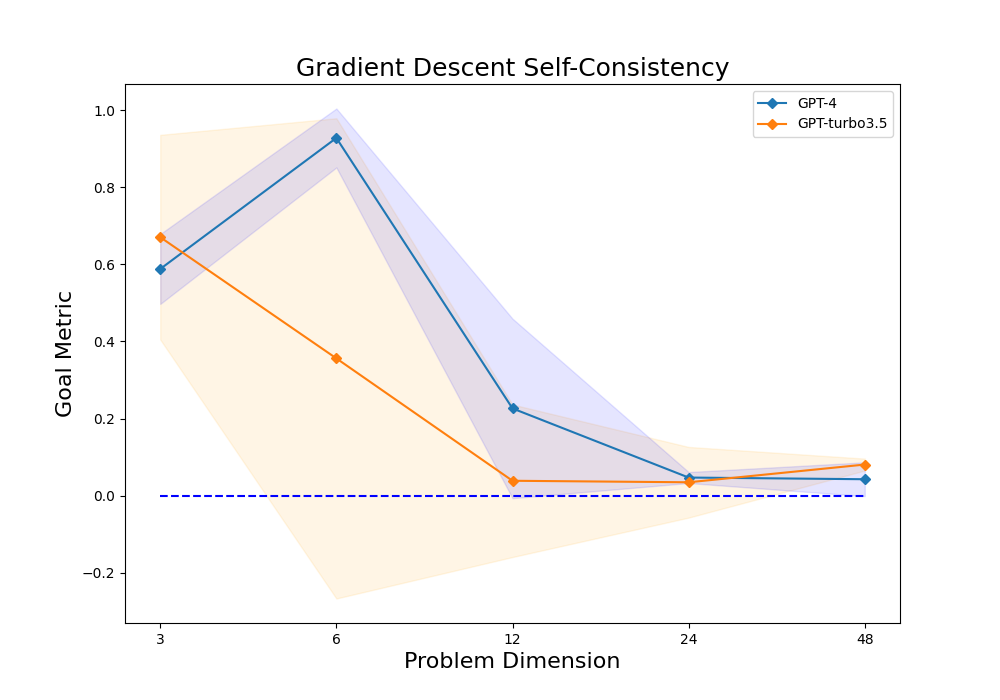}
    \label{fig:image9}
  \end{minipage}
  \hfill
  \begin{minipage}[h]{1.0\linewidth}
    \centering
    \includegraphics[width=\textwidth]{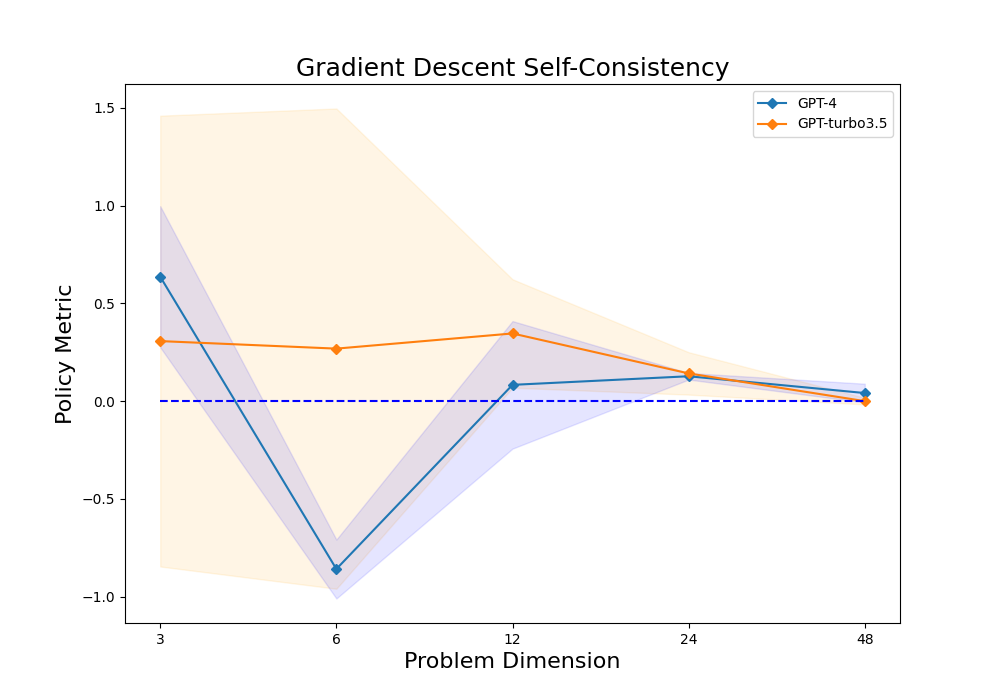}
    \label{fig:image10}
  \end{minipage}
\caption{The confidence intervals for both the policy and goal metrics of GPT-4 narrow, indicating improved stability. A negative policy metric with a high goal metric signifies significant outperformance of the ground truth model with six-dimensional instances.}
\label{gdsc}
\end{figure}

% \subsection{Ablation Study}
{\bfseries Self-consistency prompting improves stability.} In the Gradient-Descent task, we employ self-consistency technique~\cite{sc}, where we conduct five repetitions for each iteration and select the solution that emerges most frequently. From Figure~\ref{gdsc}, we can see that GPT-4 performances increase largely, and the confidence interval for both the policy-metric and goal-metric narrows, indicating improved stability and reliability. Nonetheless, this approach does not yield favorable outcomes when applied to GPT-turbo-3.5. This suggests the need for further investigation within the realm of variance reduction.

\section{Conclusion and Future Directions}
\label{sec:conclude}
In this paper, we present our in-depth examination of assessing Large Language Models within the realm of optimization, where LLM progressively generates new solutions to optimize an objective function. We investigate LLMs' performance across four optimization tasks that necessitate their comprehension of algorithmic instructions and their ability to generate new solutions based on previous solutions and their corresponding values.

Our evaluation shows that LLMs showcase optimization prowess across diverse domains. Among the four tasks we examined, LLMs exhibit their greatest strengths in the Gradient-Descent task, displaying remarkable proficiency in this area. However, they encounter more pronounced difficulties in the meta-heuristics task, where they must adhere to predefined rules that they have not encountered previously. Furthermore, LLMs demonstrate impressive skills in the grid search task, showcasing their ability to conduct exhaustive searches effectively. In the Black-Box task, LLMs excel, particularly when dealing with limited sample sizes, suggesting inherent optimization abilities within them.

We also consolidate several crucial insights derived from our experimental results and subject them to analysis. We find that pretrained knowledge dominates the optimization capability of LLMs, while they also possess a richer solution space in small-scale problems. Furthermore, we elaborate on the potential of LLMs as hybrid optimizers. These insights and analyses unveil a host of unresolved questions that warrant further research.

% \begin{acks}
% \end{acks}


% \clearpage

\bibliographystyle{ACM-Reference-Format}
\bibliography{sample-base}


\begin{thebibliography}{9}

%%% the Web address.

\ifx \showCODEN    \undefined \def \showCODEN     #1{\unskip}     \fi
\ifx \showDOI      \undefined \def \showDOI       #1{#1}\fi
\ifx \showISBNx    \undefined \def \showISBNx     #1{\unskip}     \fi
\ifx \showISBNxiii \undefined \def \showISBNxiii  #1{\unskip}     \fi
\ifx \showISSN     \undefined \def \showISSN      #1{\unskip}     \fi
\ifx \showLCCN     \undefined \def \showLCCN      #1{\unskip}     \fi
\ifx \shownote     \undefined \def \shownote      #1{#1}          \fi
\ifx \showarticletitle \undefined \def \showarticletitle #1{#1}   \fi
\ifx \showURL      \undefined \def \showURL       {\relax}        \fi
% invisible to TeX
\providecommand\bibfield[2]{#2}
\providecommand\bibinfo[2]{#2}
\providecommand\natexlab[1]{#1}
\providecommand\showeprint[2][]{arXiv:#2}

\bibitem[Chen et~al\mbox{.}(2022)]%
        {llm_knowledge}
\bibfield{author}{\bibinfo{person}{Hung-Ting Chen}, \bibinfo{person}{Michael Zhang}, {and} \bibinfo{person}{Eunsol Choi}.} \bibinfo{year}{2022}\natexlab{}.
\newblock \showarticletitle{Rich Knowledge Sources Bring Complex Knowledge Conflicts: Recalibrating Models to Reflect Conflicting Evidence}. In \bibinfo{booktitle}{\emph{Proceedings of the 2022 Conference on Empirical Methods in Natural Language Processing, pages 2292–2307, Abu Dhabi, United Arab Emirates, Association for Computational Linguistics.}}
\newblock


\bibitem[Guo et~al\mbox{.}(2023)]%
        {pw}
\bibfield{author}{\bibinfo{person}{Qingyan Guo}, \bibinfo{person}{Rui Wang}, \bibinfo{person}{Junliang Guo}, \bibinfo{person}{Bei Li}, \bibinfo{person}{Kaitao Song}, \bibinfo{person}{Xu Tan}, \bibinfo{person}{Guoqing Liu}, \bibinfo{person}{Jiang Bian}, {and} \bibinfo{person}{Yujiu Yang}.} \bibinfo{year}{2023}\natexlab{}.
\newblock \showarticletitle{Connecting Large Language Models with Evolutionary Algorithms Yields Powerful Prompt Optimizers}.
\newblock \bibinfo{journal}{\emph{arXiv:2309.08532}}.
\newblock


\bibitem[Pagnoni et~al\mbox{.}(2021)]%
        {llm_summary}
\bibfield{author}{\bibinfo{person}{Artidoro Pagnoni}, \bibinfo{person}{Vidhisha Balachandran}, {and} \bibinfo{person}{Yulia Tsvetkov}.} \bibinfo{year}{2021}\natexlab{}.
\newblock \showarticletitle{Understanding factuality in abstractive summarization with FRANK: A benchmark for factuality metrics}. In \bibinfo{booktitle}{\emph{Proceedings of the 2021 Conference of the North American Chapter of the Association for Computational Linguistics: Human Language Technologies, pages 4812–4829, Online. Association for Computational Linguistics}}.
\newblock


\bibitem[Renda et~al\mbox{.}(2023)]%
        {llm_rand}
\bibfield{author}{\bibinfo{person}{Alex Renda}, \bibinfo{person}{Aspen Hopkins}, {and} \bibinfo{person}{Michael Carbin}.} \bibinfo{year}{2023}\natexlab{}.
\newblock \showarticletitle{Can LLMs Generate Random Numbers? EvaluatingLLM Sampling in Controlled Domains}. In \bibinfo{booktitle}{\emph{ICML 2023 Workshop: Sampling and Optimization in Discrete Space}}.
\newblock


\bibitem[Shinn et~al\mbox{.}(2023)]%
        {rl}
\bibfield{author}{\bibinfo{person}{Noah Shinn}, \bibinfo{person}{Federico Cassano}, \bibinfo{person}{Beck Labash}, \bibinfo{person}{Ashwin Gopinath}, \bibinfo{person}{Karthik Narasimhan}, {and} \bibinfo{person}{Shunyu Yao}.} \bibinfo{year}{2023}\natexlab{}.
\newblock \showarticletitle{Reflexion: Language Agents with Verbal Reinforcement Learning}.
\newblock \bibinfo{journal}{\emph{arXiv:2303.11366}}.
\newblock


\bibitem[Wang et~al\mbox{.}(2022)]%
        {sc}
\bibfield{author}{\bibinfo{person}{X. Wang}, \bibinfo{person}{J. Wei}, \bibinfo{person}{D. Schuurmans}, \bibinfo{person}{Q. Le}, \bibinfo{person}{E. Chi}, {and} \bibinfo{person}{D Zhou}.} \bibinfo{year}{2022}\natexlab{}.
\newblock \showarticletitle{Self-consistency improves chain of thought reasoning in language models}.
\newblock \bibinfo{journal}{\emph{arXiv preprint arXiv:2203.11171}}.
\newblock


\bibitem[Wei et~al\mbox{.}(2022)]%
        {cot}
\bibfield{author}{\bibinfo{person}{J. Wei}, \bibinfo{person}{X. Wang}, \bibinfo{person}{D. Schuurmans}, \bibinfo{person}{M. Bosma}, \bibinfo{person}{E. Chi}, \bibinfo{person}{Q. Le}, {and} \bibinfo{person}{D Zhou}.} \bibinfo{year}{2022}\natexlab{}.
\newblock \showarticletitle{Chain of thought prompting elicits reasoning in large language models}.
\newblock \bibinfo{journal}{\emph{arXiv preprint arXiv:2201.11903}}.
\newblock


\bibitem[Yang et~al\mbox{.}(2023)]%
        {llm_as_opt}
\bibfield{author}{\bibinfo{person}{Chengrun Yang}, \bibinfo{person}{Xuezhi Wang}, \bibinfo{person}{Yifeng Lu}, \bibinfo{person}{Hanxiao Liu}, \bibinfo{person}{Quoc~V. Le}, \bibinfo{person}{Denny Zhou}, {and} \bibinfo{person}{Xinyun Chen}.} \bibinfo{year}{2023}\natexlab{}.
\newblock \showarticletitle{Large Language Models as Optimizers}.
\newblock \bibinfo{journal}{\emph{arXiv preprint arXiv:2309.03409}}.
\newblock


\bibitem[Zhou et~al\mbox{.}(2023)]%
        {llm_aware}
\bibfield{author}{\bibinfo{person}{Wenxuan Zhou}, \bibinfo{person}{Sheng Zhang}, \bibinfo{person}{Hoifung Poon}, {and} \bibinfo{person}{Muhao Chen}.} \bibinfo{year}{2023}\natexlab{}.
\newblock \showarticletitle{Context-faithful prompting for large language models}. In \bibinfo{booktitle}{\emph{ArXiv, abs/2303.11315}}.
\newblock


\end{thebibliography}


\clearpage
\appendix
\onecolumn

\section{Prompt Templates}
\label{app:prompt}

\begin{figure}[h]
\begin{tcolorbox}[width=1.0\linewidth, halign=left, colframe=black, colback=white, boxsep=0.01mm, arc=1.5mm, left=2mm, right=2mm, boxrule=0.5pt]\footnotesize

\textbf{User Prompt:}\\ 
\textbf{Q :} \\
Given the data points (y1, y2, ...) = $\{$data$\}$, what is the MSE loss function with respect to the \^ys for a hypothetical set of predicted \^ys values?\\
\textbf{A :}\\ 
The MSE loss function for the given data points (y1, y2, ...) = $\{$data$\}$ with respect to \^ys is:... \\

\end{tcolorbox}
\caption{Example prompt for getting objective function.}
\label{fig:prompt-objfunc}
\end{figure}

\begin{figure}[h]
\begin{tcolorbox}[width=1.0\linewidth, halign=left, colframe=black, colback=white, boxsep=0.01mm, arc=1.5mm, left=2mm, right=2mm, boxrule=0.5pt]\footnotesize

\textbf{User Prompt:}\\ 
\textbf{Q :}\\ 
Please minimize the loss function using gradient descent with learning rate of 0.1 at point (\^y1, \^y2, \^y3, .....) = $\{$point$\}$. What is the point we eventually end up after one update? Your answer includes two parts an explanation with calculation and a short answer of result.\\
\textbf{A :}\\
Explanation : Lets think step by step ...\\
Short Answer: After calculation, the next update point is (\^y$1_{new}$, \^y$2_{new}$, \^y$3_{new}$, .....) = ...\\
\end{tcolorbox}
\caption{Example prompt for Gradient-Descent.}
\label{fig:prompt-summarization}
\end{figure}

\begin{figure}[h]
\begin{tcolorbox}[width=1.0\linewidth, halign=left, colframe=black, colback=white, boxsep=0.01mm, arc=1.5mm, left=2mm, right=2mm, boxrule=0.5pt]\footnotesize

\textbf{User Prompt :}\\ 
\textbf{Q :}\\ 
I want to do grid search on the \^ys and the range of them are the integers of $\{$low\_bound$\}$ to $\{$high\_bound$\}$. Generate all possible combinations of \^ys values from the specify range. \\
What are the combinations? Your answer includes two parts an explanation with calculation and a list containing all the combinations.\\
\textbf{A :}\\ 
Explanation : Lets think step by step ...\\
List : $[$write all the combinations here$]$\\

\end{tcolorbox}
\caption{Example prompt for Grid-Search (Create Grid Points)}
\label{fig:prompt-gridc}
\end{figure}

\begin{figure}[h]
\begin{tcolorbox}[width=1.0\linewidth, halign=left, colframe=black, colback=white, boxsep=0.01mm, arc=1.5mm, left=2mm, right=2mm, boxrule=0.5pt]\footnotesize

\textbf{User Prompt :}\\ 
\textbf{Q :}\\ 
For every combinations of \^ys, calculate its MSE loss. Which combination has the smallest MSE loss? Your answer includes two parts an explanation with calculation and a list containing the combination with the smallest MSE loss.\\
\textbf{A :}\\ Explanation : Lets think step by step\\
List : [write the combination with smallest MSE loss]\\

\end{tcolorbox}
\caption{Example prompt for Grid-Search (Select)}
\label{fig:prompt-grids}
\end{figure}

\begin{figure}[h]
\begin{tcolorbox}[width=1.0\linewidth, halign=left, colframe=black, colback=white, boxsep=0.01mm, arc=1.5mm, left=2mm, right=2mm, boxrule=0.5pt]\footnotesize

\textbf{User Prompt :}\\ 
\textbf{Q :}\\ 
I want to minimize the loss function using hill climbing. Generate neighboring solutions by either adding 1 or minus 1 to a specific element in the current solution. The current solution is {solution}. Your answer includes two parts an explanation with calculation and a list containing all neighbor solutions(eg. [(\^y1, \^y2,....), (\^y1, \^y2,....), ...]).\\
\textbf{A :}\\ 
Explanation : Let's think step by step ...\\
List : [write neighbor solutions here]\\

\end{tcolorbox}
\caption{Example prompt for Hill-Climbing Prompt (Generate Neighbors)}
\label{fig:prompt-hillg}
\end{figure}

\begin{figure}[h]
\begin{tcolorbox}[width=1.0\linewidth, halign=left, colframe=black, colback=white, boxsep=0.01mm, arc=1.5mm, left=2mm, right=2mm, boxrule=0.5pt]\footnotesize

\textbf{User Prompt :}\\ 
\textbf{Q :}\\ 
For every neighbor solution, calculate its MSE loss. Which neighbor has the smallest MSE loss? Your answer includes two parts an explanation with calculation and a list containing the smallest MSE loss neighbor solution(eg. [(\^y1, \^y2,....)]).\\
\textbf{A :}\\ Explanation : Lets think step by step ...\\
List : [write best neighbor solution here]\\

\end{tcolorbox}
\caption{Example prompt for Hill-Climbing Prompt (Select)}
\label{fig:prompt-hills}
\end{figure}

\begin{figure}[h]
\begin{tcolorbox}[width=1.0\linewidth, halign=left, colframe=black, colback=white, boxsep=0.01mm, arc=1.5mm, left=2mm, right=2mm, boxrule=0.5pt]\footnotesize

\textbf{User Prompt :}\\ 
\textbf{Q :}\\ 
You want to minimize an unknown MSE loss function by guessing the values of the \^ys. When you guess, you should take consider of the past guessing result so that your new guess will have smaller loss than the past results. Pass guessing result are $\{$pass\_result$\}$. Base on the previous guesses, what is your next guess? \\
\textbf{A :}\\ (\^y1, \^y2,....) = [your answer]\\

\end{tcolorbox}
\caption{Example prompt for Black-Box}
\label{fig:prompt-bb}
\end{figure}

\begin{figure}[h]
\begin{tcolorbox}[width=1.0\linewidth, halign=left, colframe=black, colback=white, boxsep=0.01mm, arc=1.5mm, left=2mm, right=2mm, boxrule=0.5pt]\footnotesize

\textbf{User Prompt :}\\ 
\textbf{*Define Loss function}\\
\textbf{User :}\\
Given the data points (y1, y2, y3) = (2,6,0), what is the MSE loss function with respect to the \^ys for a hypothetical set of predicted \^ys values?\\
\textbf{LLM :}\\ 
The MSE loss function for the given data points (y1, y2, y3) = (2,6,0) with respect to \^ys is: $\frac{1}{3}[(\hat{y}_1-2)^2 + (\hat{y}_2-6)^2 + (\hat{y}_3-0)^2]$ \\
\textbf{*First Iteration}\\
\textbf{User :}\\ Please minimize the loss function using gradient descent with learning rate of 0.1 at point (\^y1, \^y2, \^y3) = (2,3,4). What is the point we eventually end up after one update? Your answer includes two parts an explanation with calculation and a short answer of result.\\
\textbf{LLM :}\\
Explanation : To minimize the Mean Squared Error (MSE) loss function using gradient descent ...\\
Short Answer: After calculation, the next update point is (\^y$1_{new}$, \^y$2_{new}$, \^y$3_{new}$) = (1.92, 3.05, 3.94)\\
\textbf{*Second Iteration}\\
\textbf{User :}\\ 
Please minimize the loss function using gradient descent with learning rate of 0.1 at point (\^y1, \^y2, \^y3) = ((1.92, 3.05, 3.94). What is the point we eventually end up after one update? Your answer includes two parts an explanation with calculation and a short answer of result.\\
\textbf{LLM :}\\
Explanation : To minimize the Mean Squared Error (MSE) loss function using gradient descent ...\\
Short Answer: After calculation, the next update point is (\^y$1_{new}$, \^y$2_{new}$, \^y$3_{new}$) = (1.81, 3.12, 3.41)...\\ 

\end{tcolorbox}
\caption{Iterative prompt for Gradient-Descent}
\label{fig:prompt-bb}
\end{figure}
\begin{figure}[h]
\begin{tcolorbox}[width=1.0\linewidth, halign=left, colframe=black, colback=white, boxsep=0.01mm, arc=1.5mm, left=2mm, right=2mm, boxrule=0.5pt]\footnotesize

\textbf{User Prompt :}\\ 
\textbf{*Define Loss function}\\
\textbf{User :}\\
Given the data points (y1, y2, y3) = (2,6,0), what is the MSE loss function with respect to the \^ys for a hypothetical set of predicted \^ys values?\\
\textbf{LLM :}\\ 
The MSE loss function for the given data points (y1, y2, y3) = (2,6,0) with respect to \^ys is: $\frac{1}{3}[(\hat{y}_1-2)^2 + (\hat{y}_2-6)^2 + (\hat{y}_3-0)^2]$ \\
\textbf{*Create Grid Points}\\
\textbf{User :}\\ I want to do grid search on the \^ys and the range of them are the integers of 2 to 3. Generate all possible combinations of \^ys values from the specify range. What are the combinations? Your answer includes two parts an explanation with calculation and a list containing all the combinations.\\
\textbf{LLM :}\\
Explanation : To generate all the grid points, you should ...\\
List : [(2, 2, 2), (2, 2, 3), (2, 3, 2), (2, 3, 3), (3, 2, 2), (3, 2, 3), (3, 3, 2), (3, 3, 3)]\\
\textbf{*Select Best}\\
\textbf{User :}\\ For every combinations of \^ys, calculate its MSE loss. Which combination has the smallest MSE loss? Your answer includes two parts an explanation with calculation and a list containing the combination with the smallest MSE loss.\\
\textbf{LLM :}\\
Explanation : To select the grid points with smallest MSE value ...\\
List : [(2, 3, 3)]...\\

\end{tcolorbox}
\caption{Iterative prompt for Grid-Search}
\label{fig:prompt-bb}
\end{figure}
\begin{figure}[h]
\begin{tcolorbox}[width=1.0\linewidth, halign=left, colframe=black, colback=white, boxsep=0.01mm, arc=1.5mm, left=2mm, right=2mm, boxrule=0.5pt]\footnotesize

\textbf{User Prompt :}\\ 
\textbf{*Define Loss function}\\
\textbf{User :}\\
Given the data points (y1, y2, y3) = (2,6,0), what is the MSE loss function with respect to the \^ys for a hypothetical set of predicted \^ys values?\\
\textbf{LLM :}\\ 
The MSE loss function for the given data points (y1, y2, y3) = (2,6,0) with respect to \^ys is: $\frac{1}{3}[(\hat{y}_1-2)^2 + (\hat{y}_2-6)^2 + (\hat{y}_3-0)^2]$ \\
\textbf{*Generate Neighbors}\\
\textbf{User :}\\ 
I want to minimize the loss function using hill climbing. Generate neighboring solutions by either adding 1 or minus 1 to a specific element in the current solution. The current solution is (10,10,10). Your answer includes two parts an explanation with calculation and a list containing all neighbor solutions(eg. [(\^y1, \^y2,....), (\^y1, \^y2,....), ...]).\\
\textbf{LLM :}\\
Explanation : To generate all the neighbors ...\\
List : [(11, 10, 10), (10, 11, 10), (10, 10, 11), (9, 10, 10), (10, 9, 10), (10, 10, 9)].\\
\textbf{*Select Best}\\
\textbf{User :}\\ 
For every neighbor solution, calculate its MSE loss. Which neighbor has the smallest MSE loss? Your answer includes two parts an explanation with calculation and a list containing the smallest MSE loss neighbor solution(eg. [(\^y1, \^y2,....)]).\\
\textbf{LLM :}\\
Explanation : To select the neighbor solution with smallest MSE value ...\\
List : [(9, 10, 10)]\\

\end{tcolorbox}
\caption{Iterative prompt for Hill-Climbing}
\label{fig:prompt-bb}
\end{figure}

\begin{figure}[h]
\begin{tcolorbox}[width=1.0\linewidth, halign=left, colframe=black, colback=white, boxsep=0.01mm, arc=1.5mm, left=2mm, right=2mm, boxrule=0.5pt]\footnotesize

\textbf{User Prompt :}\\ 
\textbf{*Define Loss function}\\
\textbf{User :}\\
Given the data points (y1, y2, y3) = (2,6,0), what is the MSE loss function with respect to the \^ys for a hypothetical set of predicted \^ys values?\\
\textbf{LLM :}\\ The MSE loss function for the given data points (y1, y2, y3) = (2,6,0) with respect to \^ys is: $\frac{1}{3}[(\hat{y}_1-2)^2 + (\hat{y}_2-6)^2 + (\hat{y}_3-0)^2]$ \\
\textbf{First Iteration}\\
\textbf{User :}\\ You want to minimize an unknown MSE loss function by guessing the values of the \^ys. When you guess, you should take consider of the past guessing result so that your new guess will have smaller loss than the past results. Pass guessing result are $f(2,4,6) = 10$, $f(1,5,0)=2$. Base on the previous guesses, what is your next guess? \\
\textbf{LLM :}\\ (\^y1, \^y2, \^y3) = [$1,3,5$]\\ 
\textbf{Second Iteration}\\
\textbf{User :}\\ You want to minimize an unknown MSE loss function by guessing the values of the \^ys. When you guess, you should take consider of the past guessing result so that your new guess will have smaller loss than the past results. Pass guessing result are $f(2,4,6) = 10$, $f(1,5,0)=2$, $f(1,3,5)=5$. Base on the previous guesses, what is your next guess? \\
\textbf{LLM :}\\ 
(\^y1, \^y2, \^y3) = [$1,1,1$]\\
\end{tcolorbox}
\caption{Iterative prompt for Black-Box Optimization}
\label{fig:prompt-bb}
\end{figure}

\clearpage

\end{document}